%% file: main_IROS.tex
\documentclass[letterpaper, 10 pt, conference]{ieeeconf} 

\IEEEoverridecommandlockouts                              

\usepackage[pagebackref,breaklinks,colorlinks]{hyperref}

\usepackage{graphicx}
\usepackage{soul}
\let\labelindent\relax
\usepackage[inline]{enumitem}
\usepackage{hyperref}
\usepackage{booktabs}
\usepackage{balance}

\input{h2t_def}
\usepackage{amssymb}
\usepackage{pifont}
\newcommand{\cmark}{\color{green}\ding{51}}%
\newcommand{\xmark}{\color{red}\ding{55}}%
\title{\LARGE \bf
KITchen: A Real-World Benchmark and Dataset for 6D Object Pose Estimation in Kitchen Environments}

\author{Abdelrahman Younes and Tamim Asfour%
\thanks{The research leading to these results has received funding from the Baden-Württemberg Ministry of Science, Research and the Arts (MWK) as part of the state's "digital@bw" digitization strategy in the context of the Real-World Lab "Robotics AI", the Carl Zeiss Foundation through the JuBot project and the German Federal Ministry of Education and Research (BMBF) under the Robotics Institute Germany (RIG). The authors are with the Institute for Anthropomatics and Robotics, Karlsruhe Institute of Technology, Karlsruhe, Germany. {\tt \{younes,asfour\}@kit.edu}}
}

\begin{document}

\maketitle
\thispagestyle{empty}
\pagestyle{empty}

\begin{abstract}

Despite the recent progress on 6D object pose estimation methods for robotic grasping, a substantial performance gap persists between the capabilities of these methods on existing datasets and their efficacy in real-world grasping and mobile manipulation tasks, particularly when robots rely solely on their monocular egocentric field of view (FOV). Existing real-world datasets primarily focus on table-top grasping scenarios, where a robot arm is placed in a fixed position and the objects are centralized within the FOV of fixed external camera(s).  
Assessing performance on such datasets may not accurately reflect the challenges encountered in everyday grasping and mobile manipulation tasks within kitchen environments such as retrieving objects from higher shelves, sinks, dishwashers, ovens, refrigerators, or microwaves.
To address this gap, we present KITchen, a novel benchmark designed specifically for estimating the 6D poses of objects located in diverse positions within kitchen settings. For this purpose, we recorded a comprehensive dataset comprising around 205k real-world RGBD images for 111 kitchen objects captured in two distinct kitchens, utilizing a humanoid robot with its egocentric perspectives. Subsequently, we developed a semi-automated annotation pipeline, to streamline the labeling process of such datasets, resulting in the generation of 2D object labels, 2D object segmentation masks, and 6D object poses with minimal human effort.
The benchmark, the dataset, and the annotation pipeline are publicly available at \url{https://kitchen-dataset.github.io/KITchen}.
\end{abstract}

\section{Introduction}\label{sec:intro}
\input{sections/1_intro}

\section{Related Work}\label{sec:relatedwork}
\input{sections/2_related_work}

\section{The KITchen Dataset}\label{sec::dataset}
\input{sections/3_dataset}

\section{The KITchen Benchmark}\label{sec:benchmark}

\input{sections/4_benchmark}

\section{Conclusion}
\input{sections/5_conclusion}

\section*{Acknowledgment}
We would like to thank Diana Burkart and Lisa Joosten for their contributions and assistance during the annotation process of the dataset.

\balance 
\bibliographystyle{IEEEtran}
\bibliography{egbib}

\end{document}

%% file: h2t_def.tex
\usepackage{comment}
\usepackage[normalem]{ulem}
\usepackage{xspace}
\usepackage{xcolor}

\definecolor{OliveGreen}{RGB}{0,200,25}
\newcommand{\red}[1]{\textcolor{red}{#1}}

\newcommand{\darkgreen}[1]{\textcolor{OliveGreen}{#1}}
\newcommand{\blue}[1]{\textcolor{blue}{#1}}
\newcommand{\orange}[1]{\textcolor{orange}{#1}}

\newcommand{\armarVI}{\mbox{ARMAR-6}\xspace}

\newcommand{\added}[1]{\darkgreen{#1}} 
\newcommand{\replaced}[2]{\red{\ifmmode\text{\sout{\ensuremath{#1}}}\else\sout{#1}\fi}\darkgreen{#2}}
\newcommand{\removed}[1]{\red{\ifmmode\text{\sout{\ensuremath{#1}}}\else\sout{#1}\fi}}
\newcommand{\remark}[1]{\blue{#1}}

\newcommand{\todo}[1]{\{\orange{---TODO--- #1}\}}

\newif\iffinal

\iffinal
	\renewcommand{\added}[1]{#1}
	\renewcommand{\replaced}[2]{#2}
	\renewcommand{\replaced}[2]{\added{#2}}
	\renewcommand{\removed}[1]{}
	\renewcommand{\remark}[1]{}
	\renewcommand{\todo}[1]{}
\fi

\newcommand{\removedfootnote}[1]{\footnote{\removed{#1}}}
\newcommand{\removedsubsection}[1]{\subsection{\texorpdfstring{\removed{#1}}{#1}}}

%% file: sections/1_intro.tex
Recent work in robot navigation in indoor environments shows remarkable advances for mobile robots to navigate towards a goal position following different modalities such as 2D points~\cite{ye2021auxiliary, datta2021integrating}, object's image~\cite{chaplot2020object, pal2021learning}, language instruction~\cite{chang2023goat, anderson2018vision}, and acoustic signals~\cite{younes2023catch, chen2022soundspaces}. However, expanding the capabilities of these robots beyond navigation to perform tasks that require physical interaction with the surrounding objects in the environments remains a harder challenge. Therefore, understanding the 3D surroundings and objects' 6D pose estimation are essential pre-tasks for any robotic grasping and manipulation task~\cite{pohl2024memory, reister2022combining, birr2024autogpt+}. 

Current advances in tackling the 6D pose estimation problem focus on developing new models and approaches~\cite{su2022zebrapose, labbe2022megapose} to achieve the best results on the BOP challenge\footnote{\url{https://bop.felk.cvut.cz}} datasets~\cite{hinterstoisser2013model,brachmann2014learning,hodan2017t, drost2017introducing, kaskman2019homebreweddb, tyree20226, xiang2017posecnn, doumanoglou2016recovering, tejani2014latent, hodan2018bop}. While this paradigm boosted the research on 6D pose estimation, however, the available real-world datasets primarily focus on serving the table-top robotic grasping setup, featuring a robotic arm fixed in a position above objects, close to them, and often the objects are centered within the robot's FOV and in some cases with multiple cameras setup~\cite{thalhammer2023challenges}.

\begin{figure}
    \centering
    \includegraphics[width=1\linewidth]{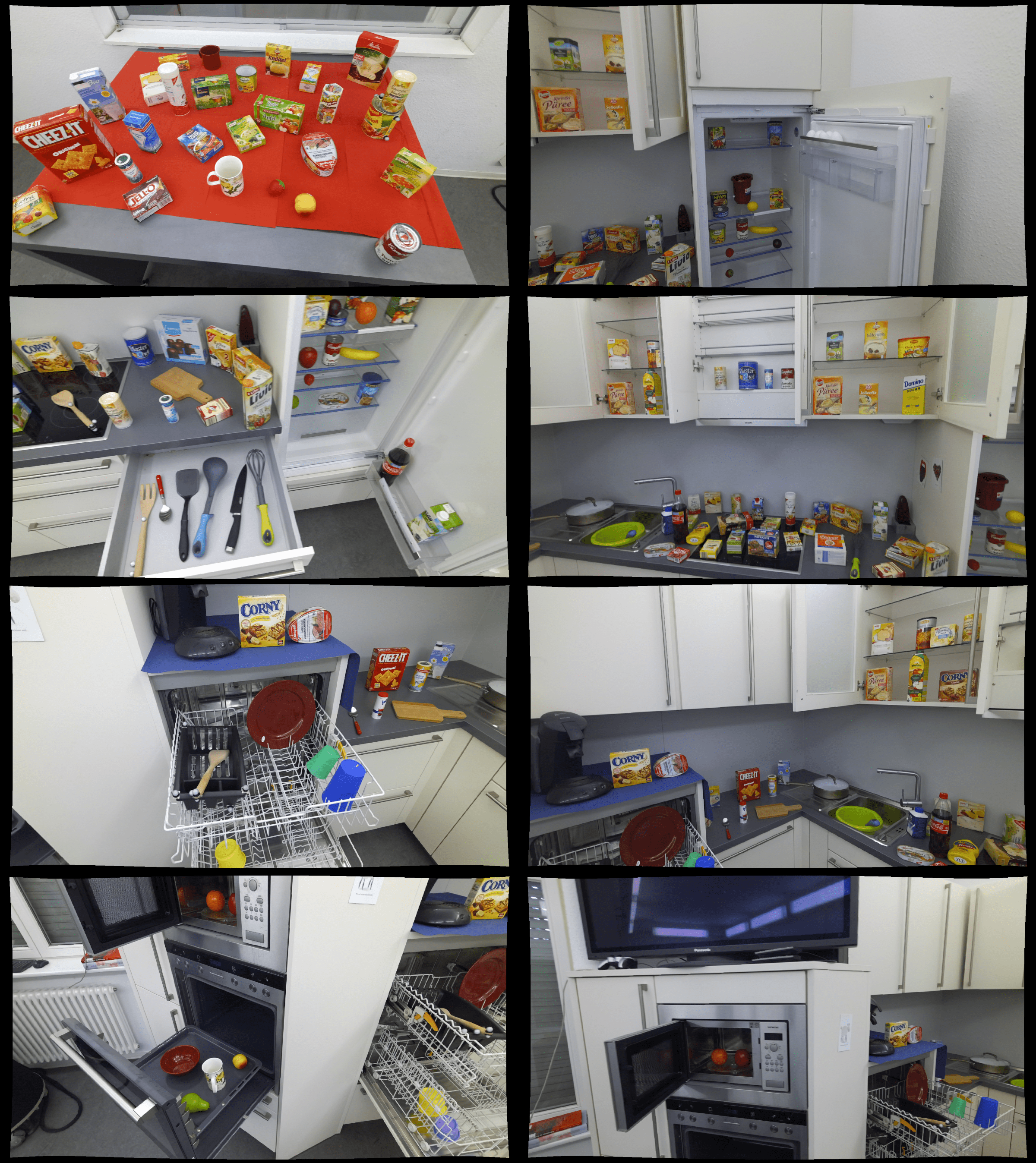}
    \caption{Challenging kitchen locations that our dataset covers in contrast with the currently available datasets. The objects are distributed across diverse locations such as fridge, drawer, sink, higher shelves, microwave, dishwasher, oven, etc.}
    \label{fig:kitchen-locations}
\end{figure}
These datasets do not cover the challenging scenarios that mobile manipulators face inside indoor environments, especially in kitchens, where objects are normally placed in different not-centered positions with respect to the robot's field of view (FOV) such as on higher shelves, inside fridges, microwaves, dishwashers or ovens or in sinks. These locations not only impose challenging 6D poses with respect to the robot's camera but also cover more diverse and challenging surroundings such as transparent shelves in the case of refrigerators, see-through shelves in the case of dishwashers, and reflective backgrounds in the case of sinks, these challenges are not covered in the currently available real-world datasets \cite{thalhammer2023challenges}. These gaps and the not-covered scenarios do not provide a reliable indication of the performance of the developed methods on these real-world datasets in the context of mobile manipulation tasks with monocular egocentric FOV.

In addition to that, the current top 10 models on the BOP leaderboard train a model for each dataset~\cite{hu2022perspective}, or even for each object~\cite{su2022zebrapose,wang2021gdr}, which makes it hard to use for robotic applications, where the robots have to deal with a large number of objects under constrained resources. Furthermore, the average inference time of these top 10 approaches is $0.0283$ frames per second ($fps$) with the best being $4.386 fps$. This makes these approaches not reliable for real-time applications, such as mobile manipulation where the 6D pose estimate is only a preliminary step of object grasping which is followed by a set of actions needed to execute the grasp such as grasp selection, motion planning, etc. 
\begin{figure}[htb]
    \centering
    \includegraphics[width=0.6\linewidth]{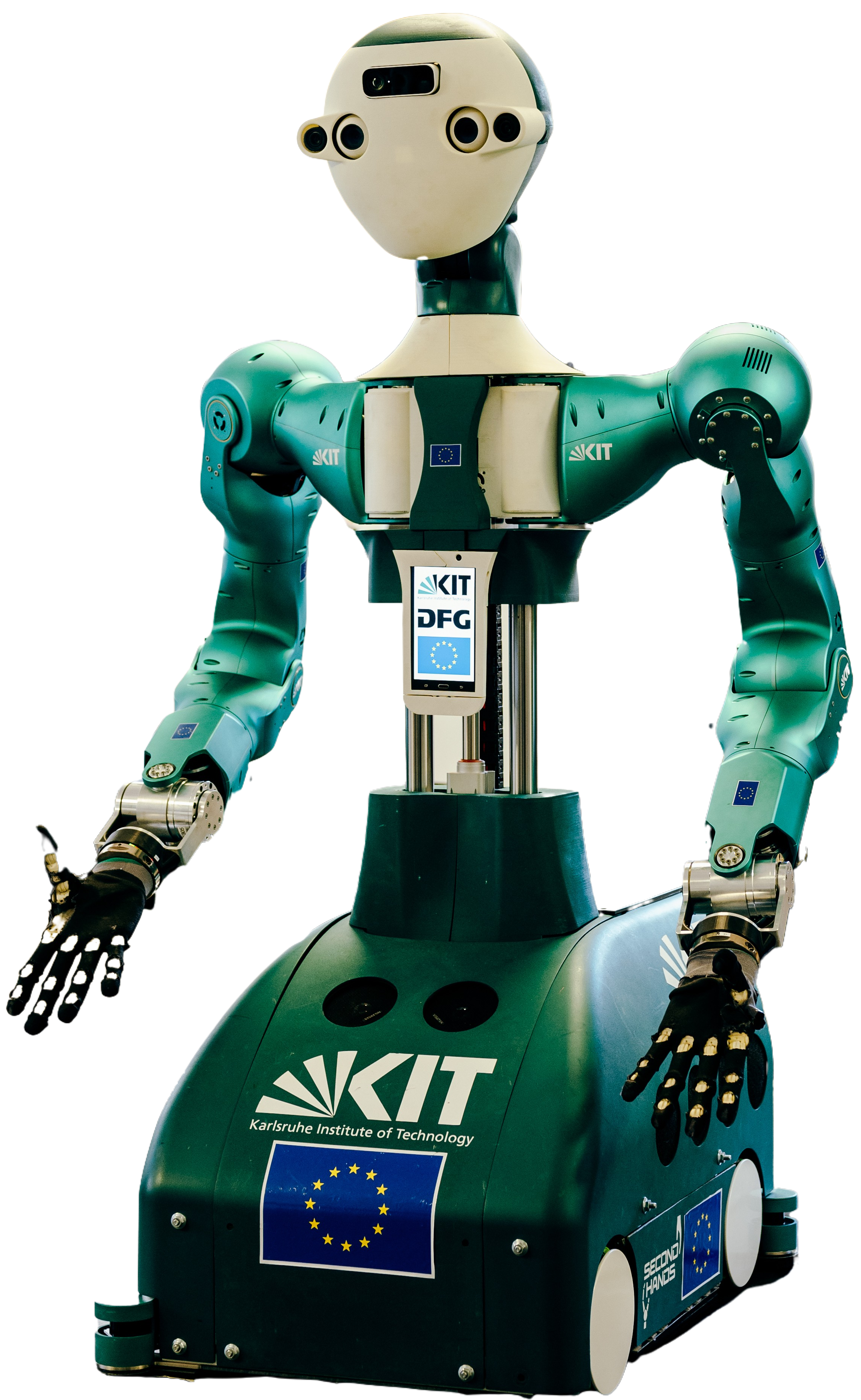}
    \caption{The humanoid robot ARMAR-6, leveraged for its adjustable torso height and various camera angles provided by its adjustable roll-yaw neck, to enrich our dataset.}
    \label{fig:ARMAR6}
\end{figure}
To overcome the limitations of the current 6D pose estimation methods, we introduce KITchen, the first-of-its-kind large-scale real-world dataset recorded using the humanoid robot \armarVI~\cite{Asfour2019} as shown in Fig.~\ref{fig:ARMAR6}, which has adjustable height and roll-yaw neck, in 2 different kitchen environments covering 111 kitchen objects from the robots' egocentric perspective to cover the objects in the challenging kitchens' locations as shown in Fig.~\ref{fig:kitchen-locations}. KITchen offers 2D bounding boxes, object segmentation, and 6D poses annotated with a semi-automated annotation pipeline to minimize the need for manual labeling. 

The main contributions of our work are: 
\begin{enumerate*}[label=(\roman*)]
    \item we introduce a large real-world annotated RGBD dataset for 111 objects with their 2D bounding boxes, segmentation masks, and 6D poses. 
    \item we propose a semi-automated annotation pipeline to annotate the objects in the dataset to facilitate the creation of more real-world datasets and make it publicly available to other researchers to create such large-scale datasets.
    \item we introduce a new benchmark and competition, where the focus is to solve the object 6D pose estimation problem depending solely on the monocular FOV of robots and limiting the submissions to approaches that offer at least $5 fps$ to encourage further work on this problem while taking into consideration real-time applicability.
\end{enumerate*}

%% file: sections/2_related_work.tex
\input{tables/datasets_comparison}
\subsection{Objects Datasets}
Current research on 6D pose estimation leverages several datasets categorized into two main groups: instance-level object datasets and category-level object datasets. Instance-level datasets offer 6D pose annotations for specific objects, serving as benchmarks for many object pose estimation methods. In contrast, category-level datasets aim to extend object pose estimation approaches to estimate the pose of different instances within the same category. In this work, we focus on instance-level object pose estimation. This subsection provides an overview of currently available real-world datasets for instance-level 6D object pose estimation. 

\textit{LineMOD (LM)}~\cite{hinterstoisser2013model} comprises 15 texture-less objects with diverse shapes, colors, and sizes. LM provides approximately 1.2K real-world test images for each object in cluttered scenes, totaling 18241 images. 
\textit{LineMOD-Occluded (LMO)}~\cite{brachmann2014learning} offers pose annotations for only eight objects from the LineMOD dataset under severe occluded conditions. 
\textit{T-LESS}~\cite{hodan2017t} consists of 30 industrial texture-less, symmetric, and similar objects with 1296 real-world images per object, totaling around 39K images. 
\textit{ITODD}~\cite{drost2017introducing} provides 6D pose annotations for 28 industrial objects with less than 1K publicly available Gray-Depth validation images. 
\textit{Homebrewed-Database (HB)}~\cite{kaskman2019homebreweddb} comprises less than 5K real-world images as validation set for 33 objects, with only 8 of them being household objects. 
\textit{HOPE}~\cite{tyree20226} consists of 28 toy grocery objects that could be utilized in kitchen environments, but it provides only 238 real-world images in 50 scenes. 
\textit{IC-BIN}~\cite{doumanoglou2016recovering} also offers only 177 real-world test images for only 3 out of its 8 objects in multi-objects cluttered scenes with heavy occlusion to be used for the BOP challenge. 
\textit{TUD-L}~\cite{hodan2018bop} provides around 11K real-world images for 3 objects not placed on tables which differs this dataset from the others. 
\textit{MP6D}~\cite{chen2022mp6d} consists of 20.1K real-world frames for 20 symmetrical specular-reflective objects in cluttered multi-object setups with occlusion.
\textit{ClearPose}~\cite{chen2022clearpose} offers about 355K real images for 63 transparent symmetrical objects in 51 cluttered scenes with diverse backgrounds and occlusion. 
\textit{YCB-video (YCB-V)}~\cite{xiang2017posecnn} provides 134K real-world images for 21 objects from the original YCB dataset~\cite{calli2015benchmarking}. 

\textit{GraspNet-1Billion}~\cite{fang2020graspnet} contains around 97.3K RGBD images for 88 objects recorded with 2 different cameras for table-top grasping scenario with one robot arm.
\textit{KIT object models database}~\cite{kasper2012kit} was originally introduced in 2012 and offers 3D CAD models for more than 100 diverse objects, the majority of which are kitchen-related groceries. However, it only offers very few images for each object, which makes it hard to use this dataset for 6D pose estimation with the current state-of-the-art (SOTA) data-driven 6D pose estimation approaches. 
\textit{KIT bimanual manipulation dataset}~\cite{KrebsMeixner2021} provides rich data for learning models of bimanual manipulation tasks from human demonstrations. It includes accurate whole-body motion data, hand configurations, and 6D object poses captured using various sensors. The dataset features 12 bimanual actions for 21 kitchen-related objects. 

An overview of available datasets for instance-level 6D pose estimation is given in~ Table~\ref{tab:dataset_comparison}. The overview highlights key metrics including the number of covered objects in the dataset, total image count, number of annotated objects per image, presence of multi-object setups, availability of multiple instances of the same objects, and whether the dataset was captured using a mobile robot's field of view.

In this work, we carefully selected 111 kitchen-related objects from the YCB, KIT object dataset, and the KIT bimanual manipulation dataset to record the first-of-its-kind large-scale real-world RGBD dataset featuring multi-objects in structured cluttered setups with diverse backgrounds and lighting conditions recorded using a humanoid robot. 

\subsection{6D Pose Estimation Methods}
The current landscape of 6D pose estimation methods is diverse, ranging from traditional techniques such as template matching~\cite{jurie2001simple, hinterstoisser2011multimodal, sundermeyer2020multi, nguyen2022templates} and correspondences with locally invariant features~\cite{collet2010efficient,collet2011moped,pauwels2015simtrack} to the current advanced deep learning  SOTA render \& compare approaches~\cite{labbe2022megapose, wen2023foundationpose}. These approaches provide the 6D poses of novel objects by rendering many views of the object during inference using its 3D CAD model and then passing these rendered views with the received cropped image of the object obtained by any 2D object detectors~\cite{tan2020efficientdet, long2020pp, zhang2021vit, li2022yolov6, wang2024yolov9} to a coarse model which classifies which rendered image best matches the input image. Finally, they pass the initial pose to a refiner network to estimate an updated 6D pose of the object. In this work, we leverage MegaPose~\cite{labbe2022megapose}, Segment Anything~\cite{kirillov2023segany}, and YOLOv8~\cite{varghese2024yolov8} to annotate our dataset.

%% file: tables/datasets_comparison.tex
\begin{table*}[htb]
    \centering
    \begin{tabular}{l|c|c|c|c|c|c}
     \toprule
        Dataset & Objects & Images & Annotated Objects/Image& Multi-object & Multi-instance  & Mobile Robot's FOV \\
        \midrule
         LineMOD/LineMOD-Occluded~\cite{hinterstoisser2013model,brachmann2014learning} & 15 & 18.2K & $\leq$ 8 &\cmark & \xmark & \xmark\\
         T-LESS~\cite{hodan2017t} & 30  & 39K & $\leq$ 10 & \cmark & \cmark& \xmark\\
         ITODD~\cite{drost2017introducing} & 28 &  1K & $\leq$ 8 &\cmark & \cmark& \xmark\\
         Homebrewed-Database~\cite{kaskman2019homebreweddb} &  33 & 5K & $\leq$ 8 &\cmark & \xmark & \xmark\\
        HOPE~\cite{tyree20226} & 28 & 238 & 5-20 &\cmark & \cmark & \xmark\\
        ICBIN~\cite{doumanoglou2016recovering}  & 3 & 177 & $\leq$ 3 &\cmark & \cmark & \xmark\\
        TUD-L~\cite{hodan2018bop} & 3 & 11K  & 1&\xmark  &\xmark & \xmark\\
        MP6D~\cite{chen2022mp6d}& 20 & 20.1K &  $\leq$ 8 & \cmark & \xmark & \xmark\\
        ClearPose~\cite{chen2022clearpose} & 63 & 355K & $\leq$ 10 & \cmark & \xmark & \xmark\\
        YCB-video (YCB-V)~\cite{xiang2017posecnn,calli2015benchmarking} & 21 & 134K & 5 & \cmark & \xmark & \xmark\\
        GraspNet-1Billion~\cite{fang2020graspnet} & 88 & 97.3K & 10 & \cmark & \xmark & \xmark\\
         KITchen (ours)& 111 & 205K & 10-50 & \cmark & \cmark & \cmark\\
         \bottomrule
    \end{tabular}
    \caption{Overview of available datasets for instance-level 6D pose estimation}
    \label{tab:dataset_comparison}
\end{table*}

%% file: sections/3_dataset.tex
\subsection{Dataset's Objects}

We aim to create a large-scale real-world dataset that covers objects that are commonly used in kitchen environments. Although some of the existing object datasets already offer objects that are commonly used in kitchens, they lack enough diverse RGBD annotated images to train on~\cite{xiang2017posecnn} or no annotated RGBD at all~\cite{calli2015benchmarking, KrebsMeixner2021, kasper2012kit, Mandery2016b}. Therefore, we decided to reuse the already available kitchen-related objects from these datasets and provide a large real-world RGBD annotated dataset for them to facilitate research on 6D pose estimation for kitchen objects. These objects vary from toy vegetables and fruits from~\cite{calli2015benchmarking} to kitchen tools such as knives, spoons, cups, mugs, bowls, cutting board, egg whisk, frying pan, plate, etc. from~\cite{calli2015benchmarking,xiang2017posecnn,KrebsMeixner2021,kasper2012kit,Mandery2016b} to kitchen groceries objects from~\cite{calli2015benchmarking,kasper2012kit}. 
\subsection{Dataset Recording}

\begin{figure*}[htb]
    \centering
    \includegraphics[width=1\linewidth]{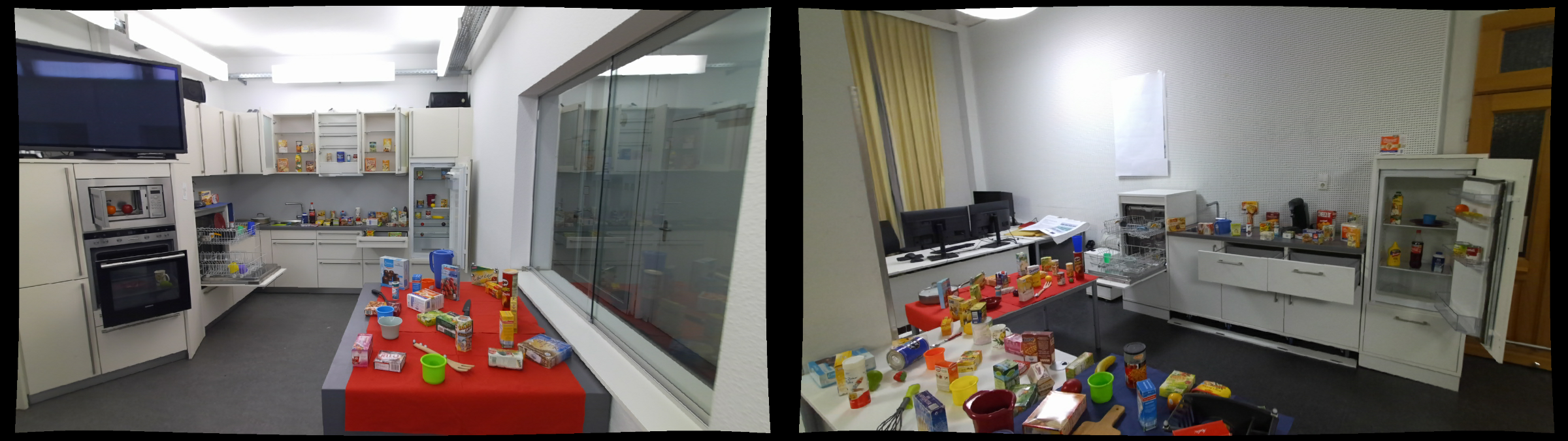}
    \caption{The two distinguished kitchens where we recorded our dataset. On the left side is the Main Kitchen while on the right side is the Mobile Kitchen.}
    \label{fig:kitchens}
\end{figure*}
\begin{figure*}[htb]
    \centering
    \includegraphics[width=1\linewidth]{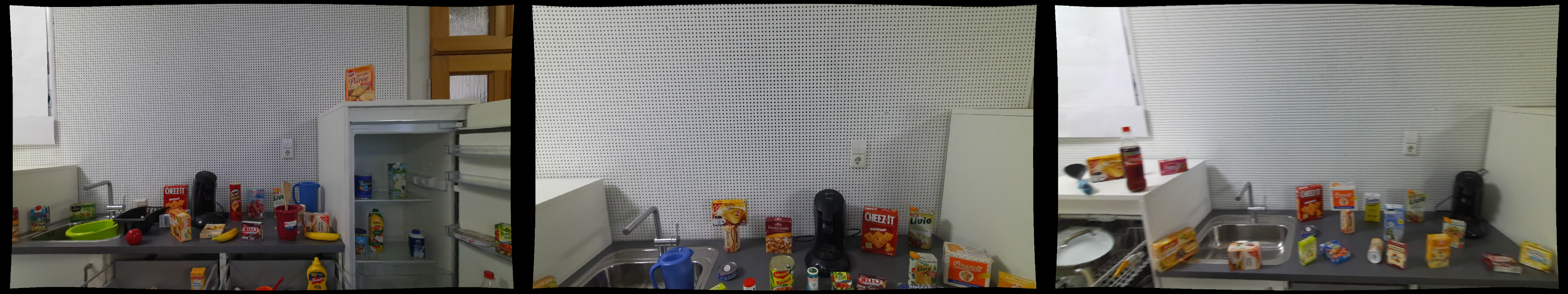}
    \caption{Diverse robot and camera heights realized through different torso positions of \armarVI. The images display heights of 145cm, 177cm, and 185cm from left to right, illustrating the varied perspectives captured in the datasets and the different placements of objects relative to the robot's field of view.}
    \label{fig:torso}
\end{figure*}
\begin{figure*}[htb]
\centering
\includegraphics[width=1\linewidth]{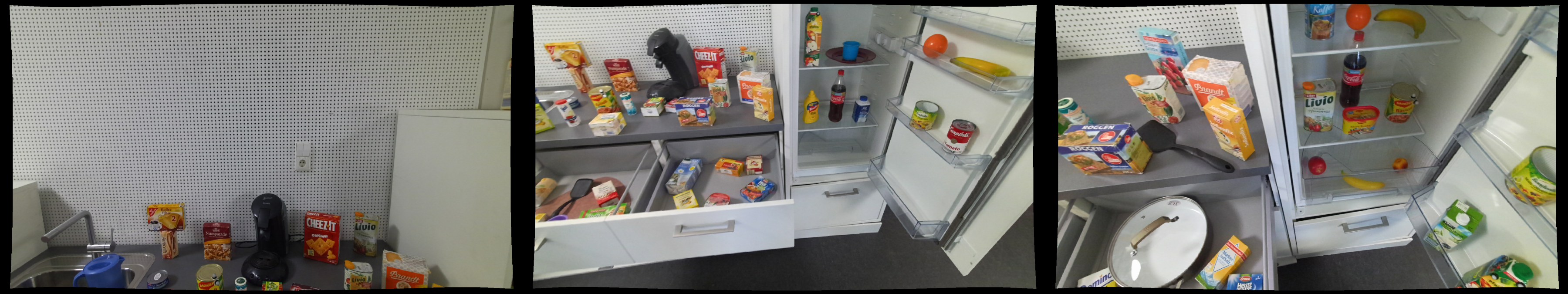}
        \caption{Variation in robot neck pitch angle. The images depict angles of 10, 37, and 49 degrees from left to right, showcasing a diverse range of perspectives.}
        \label{fig:camera-angles}
    \end{figure*}

We recorded the dataset using our humanoid robot \armarVI~\cite{asfour2019armar} inside two distinct kitchen environments as seen in Fig.~\ref{fig:kitchens} the first kitchen, referred to as the \textit{Main Kitchen}, includes typical kitchen appliances such as a fridge, counter with drawers, table, sink, microwave, dishwasher, and oven.  The second kitchen, named \textit{Mobile Kitchen}, features a counter with drawers, sink, dishwasher, fridge, and three tables. To enhance diversity, we utilized four different table-top colors (red, white, gray, and blue) and varied the camera's heights (150cm, 177cm, and 185cm) using \armarVI's torso as shown in Fig.~\ref{fig:torso}. Additionally, we recorded data under three different pitch angles (10 degrees, 37 degrees, and 49 degrees down) and six different lighting conditions as shown in Fig.~\ref{fig:camera-angles}. We shuffled the objects with each change of lighting, camera's height, or camera's angle to enrich the diversity of the recorded scenes. To avoid similar and repetitive frames, we limited our recording to 5 $fps$. To the best of our knowledge, this is the first of its kind dataset that covers this amount of different robots' fields of view.
\subsection{Annotation Pipeline}\label{sec::annotation}

\begin{figure*}[htb]
    \centering
    \includegraphics[width=1\linewidth]{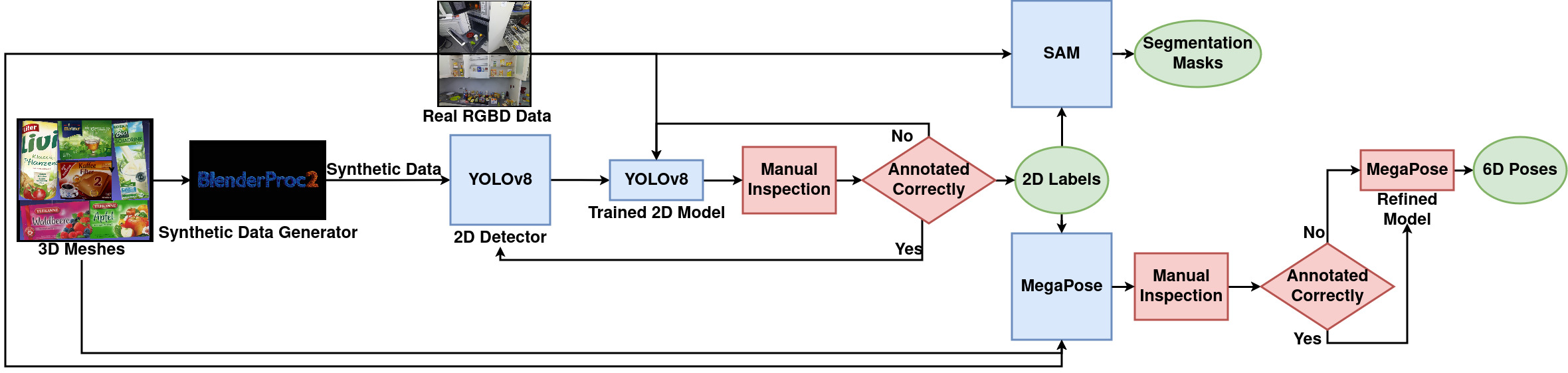}
    \caption{Our proposed annotation pipeline. The pipeline starts with 3D meshes of dataset objects as input, which are processed by BlenderProc2 to generate synthetic data with 2D bounding boxes. This annotated 2D data is used to train a YOLOv8 2D object detector. Subsequently, real-world recorded data is fed into the trained model, and the output is manually inspected for correct and incorrect labeling. The correctly labeled images are used for model refinement, which is then validated on the incorrectly labeled images. This iterative process continues until all images are correctly labeled. The correctly labeled images are then passed to Segment Anything (SAM) to generate masks. Finally, the images, along with the 2D labels and 3D meshes, are fed into MegaPose to generate 6D poses for detected objects. Manual inspection of poses is performed using contour and mesh overlay images, and corrected annotations are used to iteratively fine-tune MegaPose until the entire dataset is accurately annotated.}
    \label{fig:annotationpipeline}
\end{figure*}

Annotating objects with their ground truth 6D poses is a labor-intensive and time-consuming task. Although some of the recently published datasets attempted to semi-automate the annotation process. For instance, GraspNet-1Billion's approach~\cite{fang2020graspnet} relies on manually annotating the first frame of each scene, then leveraging recorded camera poses to calculate the objects' poses in the following frames. However, this method was not optimal for our dataset, as KITchen has many more diverse scenes per kitchen compared to the simple setup used in GraspNet-1Billion, resulting in significantly more effort required for manual annotation. Another attempt to semi-automate the annotation process was presented by HANDAL~\cite{guo2023handal},  but their approach assumes a single object in the scene, making it unsuitable for our dataset, which contains $10-50$ objects in each scene. To overcome the limitations of existing approaches and streamline the annotation process, we propose a semi-automated annotation pipeline. This pipeline generates three types of annotations: 2D object bounding boxes, 2D segmentation masks, and 6D poses, see Fig.~\ref{fig:annotationpipeline}.
\subsubsection{2D Objects Bounding Boxes Annotation}
The pipeline starts by receiving the collected 3D CAD object models for the dataset, then it generates around 100K annotated photo-realistic synthetic RGBD images with 2D bounding boxes using BlenderProc2~\cite{Denninger2023}. These synthetic images are used to finetune a pretrained YOLOv8 model~\cite{varghese2024yolov8} for 2D object detection. Subsequently, the trained model is applied to our real-world data, and manually classified images are inspected to distinguish correctly labeled ones. The model is then fine-tuned iteratively until all real-world data is accurately labeled with 2D object labels. 

\subsubsection{2D Objects Segmentation Masks}
For segmenting the objects and producing the 2D segmentation masks, we leverage Segment Anything~\cite{kirillov2023segany}, by passing the images as well as the 2D bounding boxes generated from the previous step.
\subsubsection{6D Object Poses}
\begin{figure*}[htb]
    \centering
    \includegraphics[width=1\linewidth]{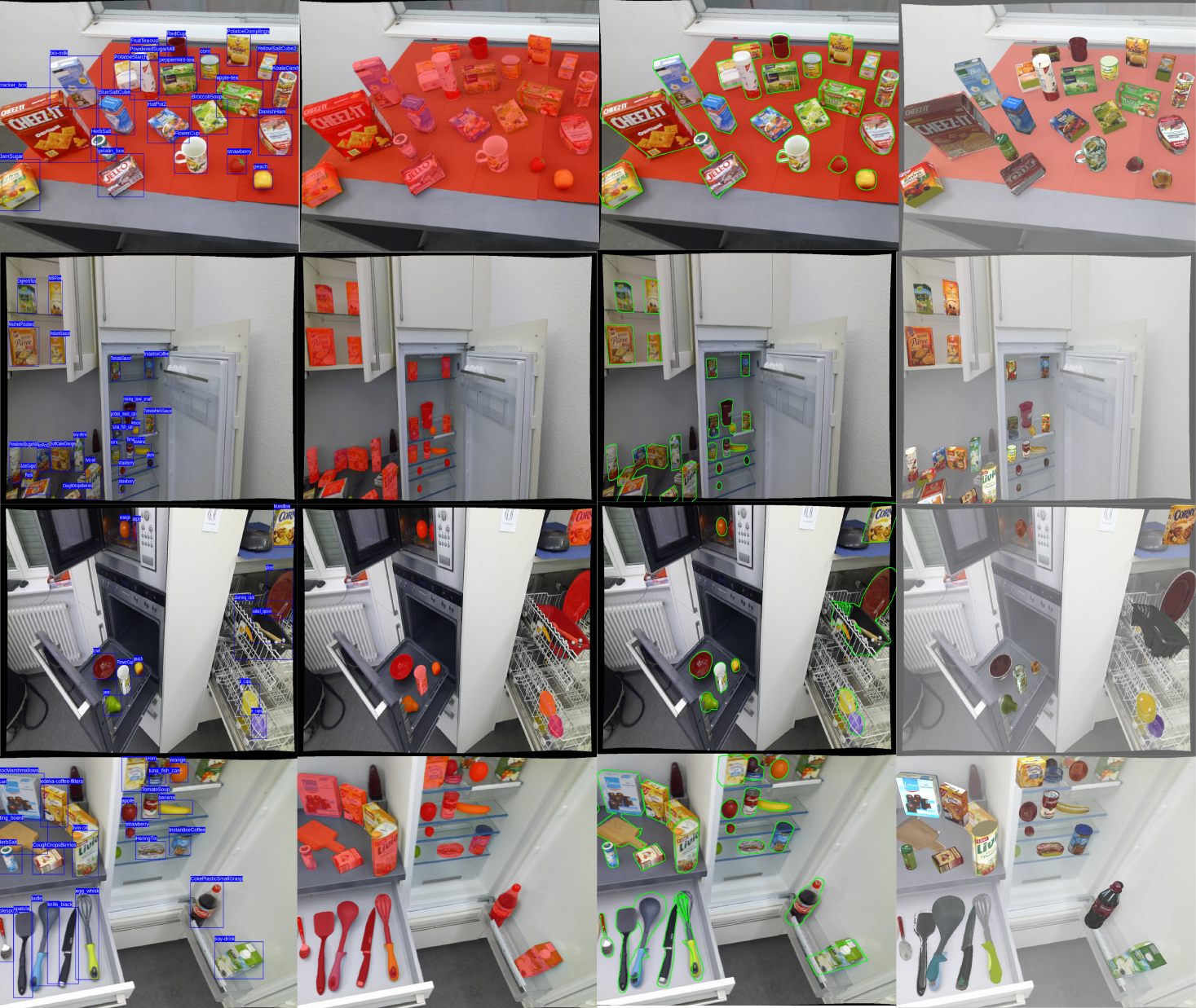}
    \includegraphics[width=1\linewidth]{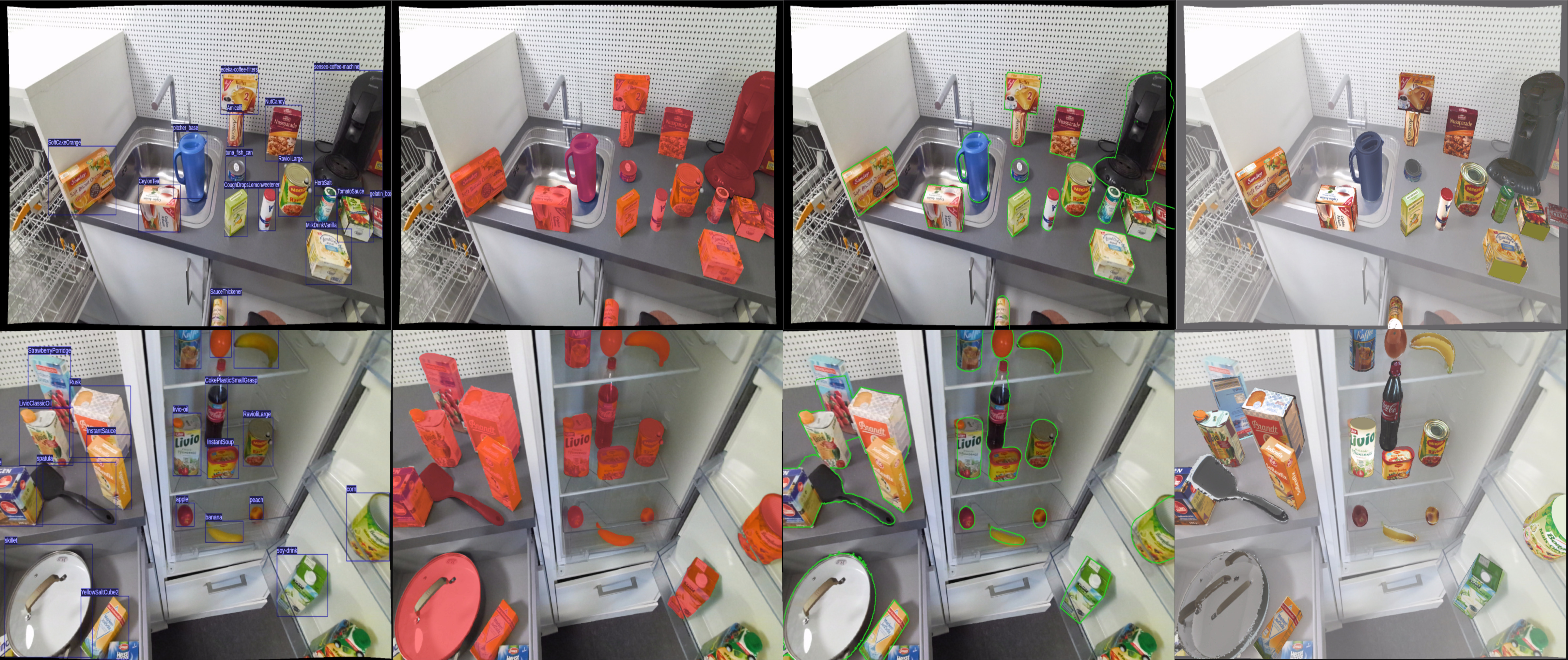}
    \caption{Examples of the results generated by our proposed annotation pipeline. Sequentially from left to right: output of the 2D detector, segmentation masks, contour overlay, and mesh overlay.}
    \label{fig:annotation-example}
\end{figure*}

To generate the 6D poses for the objects in the images, we pass the 2D bounding boxes which are generated using the fine-tuned YOLOv8 object detection model alongside the 3D CAD models of the detected objects with the input image into MegaPose~\cite{labbe2022megapose}. The output 6D poses are used to overlay contours and meshes on the images for manual inspection. The MegaPose model is fine-tuned with corrected labeled data iteratively until all data are accurately annotated.
The entire annotation pipeline is illustrated in Fig.~\ref{fig:annotationpipeline} and several illustrative examples of the output of each step are demonstrated in Fig.~\ref{fig:annotation-example}.

\subsection{Comparison to Existing Datasets}
When compared to currently available datasets, the KITchen dataset stands out in several key aspects. With a diverse collection of 111 objects, our dataset offers a significantly wider range than the average number of objects found in existing datasets, surpassing the average by a factor of four. This expansive variety is crucial for training robust pose estimation models capable of handling a multitude of real-world scenarios. Moreover, the KITchen dataset offers a total of 205K RGBD images. This surpasses the average number of annotated images in existing datasets covered in Table~\ref{tab:dataset_comparison} by over threefold, providing more data for training and evaluation purposes. Furthermore, our dataset has a remarkably larger number of annotated objects per image compared to the existing datasets with an unprecedented number of objects reaching 50 per image. This exceeds any available dataset by a significant margin, enabling more comprehensive analysis and training of instance-level 6D pose estimation models. Additionally, the KITchen dataset is unique in its capture methodology. It is the only dataset to have been recorded using the field of view of a humanoid robot with adjustable heights, camera angles, and lighting conditions. Unlike existing datasets that predominantly focus on tabletop scenes, our dataset features challenging locations within kitchen environments including refrigerators, ovens, sinks, higher shelves, microwaves, and 
dishwashers, offering a broader scope of real-world scenarios for pose estimation research. 
An overview of the dataset comparison is given in Table~\ref{tab:dataset_comparison}.

%% file: sections/4_benchmark.tex
Our proposed KITchen benchmark aims to encourage researchers in both computer vision and robotics to test their developed methods on a diverse and challenging multi-object dataset while considering the resource constraints of robots. To this end, we impose specific guidelines for leaderboard submissions to ensure practical applicability. Specifically, submissions must utilize a single model for all objects and maintain a minimum processing frequency of 5$fps$ during inference. The above conditions enhance the likelihood of the applicability of these methods in robotics. Aligning these criteria with those of the BOP Benchmark~\cite{hodavn2020bop}, we observe remarkable differences. Among the top 10 methods on the leaderboard, only two meet to the requirement of utilizing a single model per dataset rather than per object. Moreover, none of these methods achieves the required performance of 5 $fps$, with the closest reaching 4.3 fps. This discrepancy underscores a critical gap between current state-of-the-art approaches and the requirements of time-critical robotics applications, as evidenced by the average processing speed of the top 10 approaches on the BOP leaderboard, which is only  0.03 fps. 
\subsection{Problem Statement}
The benchmark is designed to address the object 6D pose estimation problem, where the model receives an image $I$ from the dataset $D$, where $D$ is a set of RGBD images. The image $I$ contains a set of objects $\{{o\}^{n}_{i=0}}$. The model has access to the $M$, where $M$ is a set of 3D meshes of all objects $O$ in the dataset $D$. The objective is to estimate the pose $P$ of all objects $\{{o\}^{n}_{i=0}}$ in each image $I$, where $P = [R, T; 0, 1]$, where $R$ is a $3 \times 3$ rotational matrix that describes the rotation of each of detected objects $\{{o\}^{n}_{i=0}}$ to the robot camera's frame and $T$ is the translation vector to the origin of robot camera's coordinate system.

\subsection{Datasets}
Our benchmark leverages the KITchen dataset introduced in Sec.~\ref{sec::dataset}. Notably, this dataset stands out as the first of its kind, captured from the perspective of a humanoid robot, and encompasses varying heights and pitch angles, making it more suited to cover robotic mobile manipulation scenarios in kitchen environments. We split the dataset to training/validation/test sets with a 70/20/10 ratio.\\
Although our benchmark primarily focuses on the KITchen dataset, we invite other robotics research groups to record datasets in kitchen environments using their own robots and leverage our proposed annotation pipeline in Sec.~\ref{sec::annotation} to annotate their data efficiently. Our vision for this benchmark extends beyond our dataset alone, we see it as a dynamic community platform where diverse research groups can collectively work to advance the field of robotic perception and pose estimation by testing their methods on a variety of datasets and providing their own datasets for other researchers to test on.

\subsection{Pose Error Calculation}
We utilize the same pose error function used by the BOP challenge~\cite{hodavn2020bop}. The estimated pose is considered correct if the pose error function $e$ calculated between the annotated pose $P$ and the estimated pose $\hat{P}$ is lower than a predefined threshold $\theta_e$, where $e$ $\in \{e_{VSD}, e_{MSSD}, e_{MSPD}\}$, where $e_{VSD}$ is the Visible Surface Discrepancy error function which focuses on the visible part of the object and evaluates poses with indistinguishable shapes as equivalent, disregarding the color information, $e_{MSSD}$ is the Maximum Symmetry-Aware Surface Distance that calculates the surface deviation between vertices in the 3D, calculating the maximum distance between model vertices is crucial to know the chance of a successful grasp, while $e_{MSPD}$ is the Maximum Symmetry-Aware Projection Distance that considers the object symmetries and calculate the difference in $X, Y$ axes which makes it suitable for methods that rely on RGB data only. Finally, the Recall is defined as the ratio of correctly estimated poses with a total pose error $e$ lower than the threshold $\theta_e$ across all objects. The Average Recall is then computed by averaging these recall values across various threshold settings.

%% file: sections/5_conclusion.tex
We introduce KITchen, a novel object 6D pose estimation benchmark tailored to tackle this task within challenging kitchen environments using only monocular vision from robots' FOV, with a specific emphasis on real-time performance. To serve this benchmark, we recorded a large-scale real-world dataset, captured from different perspectives of a humanoid robot, featuring multi-objects in structured cluttered scenes in two distinct kitchen environments with diverse lighting conditions. Lastly, we proposed a semi-automated annotation pipeline aimed at streamlining the annotation of such datasets while minimizing manual human effort. We envision our benchmark to promote the development of novel approaches to solve the 6D pose problem on resource-constrained platforms, with an emphasis on real-time and real-world applicability.